# KUKA Sunrise Toolbox: Interfacing Collaborative Robots with MATLAB

Mohammad Safeea and Pedro Neto

*Abstract*—Collaborative robots are increasingly present in our lives. The KUKA LBR iiwa equipped with the KUKA Sunrise.OS controller is a good example of a collaborative/sensitive robot. This paper presents a MATLAB Toolbox, the KUKA Sunrise Toolbox (KST), to interface KUKA Sunrise.OS using MATLAB. The KST contains functionalities for networking, real-time control, point-to-point motion, setters and getters of parameters and physical interaction. KST includes more than 50 functions and runs on a remote computer connected with the KUKA Sunrise controller via transmission control Protocol/Internet Protocol (TCP/IP). The KST potentialities are demonstrated in three use cases.

*Index Terms*—KUKA Sunrise Toolbox, MATLAB, KUKA LBR iiwa, KUKA Sunrise.OS, collaborative robotics.

## I. INTRODUCTION

### A. Motivation and Related Work

MATLAB is a high-level programming language, extensively used in engineering, research and education. It integrates computation, visualization and programming. MATLAB popularity is owed to its simplicity since programming commands are expressed in familiar mathematical notation. In addition, the software can be extended easily with extra functionalities by introducing Toolboxes.

In the robotics field several MATLAB Toolboxes had been introduced. One of the most popular is the Robotics Toolbox for MATLAB [1], [2]. This Toolbox includes functionalities for robotic manipulators, such as homogeneous transformations, direct and inverse kinematics, direct and inverse dynamics, and trajectory generation. The Dynamics simulation toolbox for industrial robot manipulators can be used for simulating robot dynamics in addition to other functionalities [3]. The DAMAROB Toolbox allows kinematic and dynamic modeling of manipulators [4]. The KUKA control toolbox is dedicated to motion control of KUKA manipulators equipped with the KRC controller [5].

Recently, KUKA lunched the LBR iiwa series of manipulators [6], sensitive robots that are programmed using the KUKA Sunrise.Workbench. From an external computer we can interface with Sunrise.OS using Robot Operating System (ROS) [7] or the fast research interface (FRI) [8]. Using ROS requires the user to have advanced technical and programming skills. The FRI is a platform for controlling the KUKA iiwa remotely from a personal computer (PC). This package is destined for researchers, people who have good technical knowledge in C++.

Mohammad Safeea is with the Department of Mechanical Engineering, University of Coimbra, Coimbra, Portugal, e-mail: ms@uc.pt.

Pedro Neto is with the Department of Mechanical Engineering, University of Coimbra, Coimbra, Portugal, e-mail: pedro.neto@dem.uc.pt.

The KUKA Sunrise.OS controller is programmed using java, allowing the internal implementation of algorithms and interfacing with external sensors. Nevertheless, our expertise indicates that the use of an external computer is advantageous in several cases:

1) To interface with multiple external devices;
2) Easy integration of external software modules;
3) Implementation of complex algorithms (requiring image processing, machine learning, etc.);
4) When the amount of computations involved is relatively high so that performance is limited by the robot controller hardware.

### B. Original Contribution

This paper presents a new MATLAB Toolbox, the KUKA Sunrise Toolbox (KST), the first Toolbox to interface KUKA Sunrise.OS that equips the KUKA LBR iiwa manipulators. The KST contains more than 50 functions divided in six categories. The advantages are multiple:

1) Easy and fast interaction with the robot from an external computer running the KST;
2) External sensors/devices are integrated in the computer and data are transmitted to/from the robot via TCP/IP;
3) Speed up the development of advanced robot applications in MATLAB. Complex algorithms can be implemented in an external computer and existing software modules/toolboxes (vision, machine learning, statistics, etc.) can be integrated;
4) The KST makes the KUKA LBR iiwa manipulators more accessible to a wide variety of people from different backgrounds, and opens the door of collaborative robotics to many potential new users for academic, educational and industrial applications.

The KST Toolbox can be freely downloaded from the Web page: https://github.com/Modi1987/KST-Kuka-Sunrise-Toolbox.

## II. KUKA SUNRISE TOOLBOX

In this section the main functions of Toolbox are illustrated with implementation examples, Table I. For the sake of convenience the functions are divided into six categories:

1) Networking – establish (and terminate) connection with the robot controller;
2) Real-time control – Activate/deactivate real-time control functionalities;
3) Point-to-point motion – Point-to-point motion in joint space and in Cartesian space;

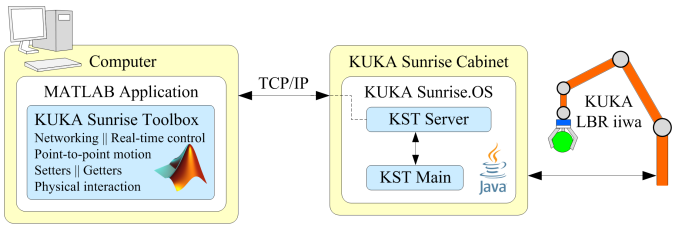

Fig. 1. Architecture and communication scheme of the KST.

4) Setters – Set parameter values in the robot controller (robot poses, LED and IO connectors);
5) Getters – Get parameter values from the robot controller (joint angles, end-effector position, end-effector orientation, force/moment acting on the end-effector, joints torques, IO connectors);
6) Physical interaction – Activate/deactivate hand-guiding and touch detection.

Figure 1 illustrates the architecture and communication scheme of the KST. The KST is running in an external/remote computer and communicates with KUKA Sunrise via TCP/IP through an Ethernet network using the X66 connector of the robot. The KST implements a TCP/IP client which communicates with the java server (KST Server and KST Main) running in the Sunrise.OS. The KST Server and KST Main are provided with the KST Toolbox.

### A. Networking

The KST provides two functionalities to establish and terminate the connection with the robot controller. The TCP/IP connection is initialized by typing:

```
>> t=net_establishConnection(ip);
```

Where `ip` is a string with the IP address of the robot controller and `t` is a TCP/IP object (to be used by other functions within the KST). The following function is used to terminate the connection:

```
>> net_turnOffServer (t);
```

### B. Real-time control

Continuous real-time control of robot motion in joint space is also allowed by the KST:

```
>> realTime_startDirectServoJoints(t);
```

Once started, we can send the target angular positions of the joints to the robot:

```
>> jPos={pi/3,0,0,-pi/2,0,pi/6,pi/2};
>> sendJointsPositions(t,jPos);
```

Where `jPos` is a cell array of dimension 1x7 containing the values of the target joints angles (KUKA LBR iiwa is a 7 DOF manipulator), in radians. A loop can be implemented for sending a stream of joint angles to the robot, so that the robot will perform the motion between the points in real-time.

To stop the real-time motion control the following function is used:

```
>> realTime_stopDirectServoJoints(t);
```

### C. Point-to-point motion

The KST point-to-point motion functionalities allow robot motion from the current configuration/end-effector pose to a defined target configuration/pose, such that continuous motion between segments cannot be achieved. The user is not required to implement the motion planning algorithms since they are integrated in the robot controller. The robot can be controlled in joint space:

```
>> jPos={pi/3,0,0,-pi/2,0,pi/6,pi/2};
>> vel=0.25;
>> movePTPJointSpace(t,jPos,vel);
```

Where `vel` represents the override velocity assuming a value from zero to one. The robot can also be controlled in Cartesian space moving along a straight line:

```
>> pos={400,0,580,-pi,0,-pi};
>> vel=50;
>> movePTPLineEEF(t,pos,vel);
```

Where `pos` is the target end-effector pose in Cartesian space, a 1x6 cell array in which the first three cells are the end-effector X,Y and Z coordinates and the last three cells are the fixed rotation angles (alpha, beta, and gamma) representing the orientation of the end-effector in the space. The variable `vel` represents the linear velocity of the end-effector in mm/sec. The KST also includes functionalities for arc and circle motion.

### D. Setters

The KST functionalities to set the values of robot parameters include robot poses, turning on and off the LED light, and to change the state of outputs of the media flange connectors of the robot. For example, to turn on the blue LED light we have:

```
>> setBlueOn(t);
```

To turn off the blue led:

```
>> setBlueOff(t);
```

### E. Getters

The KST provides functionalities to acquire various internal parameters of the robot such as joint angles, end-effector pose (position and orientation), force/moment acting on the end-effector, joints torques and the values of the IO connector inputs. For example, to acquire the joint angles of the robot, the following function is utilized:

TABLE I
LIST OF KST FUNCTIONALITIES DIVIDED BY CATEGORY.

| Category | Function | Description |
| --- | --- | --- |
| **Networking** | `net_establishConnection` | Connect to KUKA Sunrise.OS |
| | `net_turnOffServer` | Terminate connection to KUKA Sunrise.OS |
| **Real-time control** | `realTime_moveOnPathInJointSpace` | Moves the robot continuously in joint space |
| | `realTime_startDirectServoJoints` | Start the direct servo |
| | `realTime_stopDirectServoJoints` | Stop the direct servo |
| **Point-to-point motion** | `movePTPJointSpace` | Moves from the current configuration to a new configuration in joint space |
| | `movePTPLineEEF` | Moves the end-effector in a straight line from the current pose to a new pose |
| | `movePTPHomeJointSpace` | Moves the robot to the home configuration |
| | `movePTPTransportPositionJointSpace` | Moves the robot to the transportation configuration |
| | `movePTPLineEefRelBase` | Moves the end-effector in a straight line path relative to base frame |
| | `movePTPLineEefRelEef` | Moves the end-effector in a straight line path relative to end-effector initial frame |
| | `movePTPCirc1OrientationInterpolation` | Moves the end-effector in arc specified by two frames |
| | `movePTPArc_AC` | Moves the end-effector in arc specified by center, normal, arc's radius and angle |
| | `movePTPArcXY_AC` | Moves the end-effector in arc in the XY plane |
| | `movePTPArcXZ_AC` | Moves the end-effector in arc in the XZ plane |
| | `movePTPArcYZ_AC` | Moves the end-effector in arc in the YZ plane |
| **Setters** | `sendEEfPositions` | Sets in memory end-effector Cartesian positions |
| | `sendJointsPositions` | Sets the joint angles to a desired value |
| | `sendJointsPositionsf` | Sets the end-effector position to a desired value |
| | `setBlueOff` | Turns on the blue LED of the pneumatic flange |
| | `setBlueOn` | Turns off the blue LED of the pneumatic flange |
| | `setPin1Off` | Sets the output of Pin1 to low level |
| | `setPin1On` | Sets the output of Pin1 to high level |
| | To turn off and on Pin2, Pin11 and Pin 12 ( `setPin2Off`, `setPin2On`, `setPin11Off`, `setPin11On`, `setPin12Off`, `setPin12On`) | |
| **Getters** | `getEEF_Force` | Returns the measured force at the end-effector flange reference frame |
| | `getEEF_Moment` | Returns the measured moments at the end-effector flange reference frame |
| | `getEEFCartesianOrientation` | Returns the orientation (fixed rotations angles) of the end-effector in radians |
| | `getEEFCartesianPosition` | Returns the position of the end-effector relative to robot base reference frame |
| | `getEEFPos` | Returns the position and orientation of the end-effector relative to robot base reference frame |
| | `getJointsExternalTorques` | Returns the robot joint torques due to external forces |
| | `getJointsMeasuredTorques` | Returns the robot joint torques measured by the torque sensors |
| | `getJointsPos` | Returns the robot joint angles in radians |
| | `getMeasuredTorqueAtJoint` | Returns the measured torque in a specific joint |
| | `getExternalTorqueAtJoint` | Returns the measured torque in a specific joint due to external forces |
| | `getEEFOrientationR` | Returns the end-effector orientation as rotation matrix |
| | `getEEFOrientationQuat` | Returns the end-effector orientation as a quaternion |
| | `getPin3State` | Returns the state of Pin3 of the pneumatic flange |
| | Returns the state of Pin4, Pin10, Pin13 and Pin16 (`getPin4State`, `getPin10State`, `getPin13State`, `getPin16State`) | |
| **Physical interaction** | `startHandGuiding` | Initializes hand-guiding functionality |
| | `performEventFunctionAtDoubleHit` | Detects double touch |
| | `eventFunctionAtDoubleHit` | Double touch event |

```
>> jPos = getJointsPos(t);
```

The variable `jPos` we get is a 1x7 cell array with the robot joint angles in radians.

*F. Physical interaction*

The physical interaction functionalities are the hand-guiding mode and the touch detection. The hand-guiding function is activated using:

```
>> startHandGuiding(t);
```

Once called the hand-guiding functionality is initiated. To perform hand-guiding operation on the robot side we have to press the flange white button to deactivate the brakes and move the robot. After we release the white button the robot stops in its current configuration. To terminate the hand-guiding mode, we have to press the green button continuously for more than 1.5 seconds (after 1.5 seconds of pressing the green button the blue LED light starts to flicker), release the green button and the hand-guiding mode is terminated.

III. APPLICATION EXAMPLES

Three application examples on a KUKA iiwa 7 R800 manipulator (Sunrise.OS 1.11.0.7) demonstrate the performance and ease of use of KST:

1) Example 1: Hand-guiding and teaching making use of getters/setters functionalities and physical interaction;
2) Example 2: The robot drawing a rectangle exploring the point-to-point motion functionalities of KST;
3) Example 3: Human-robot collision avoidance requiring continuous motion functionalities in real-time.

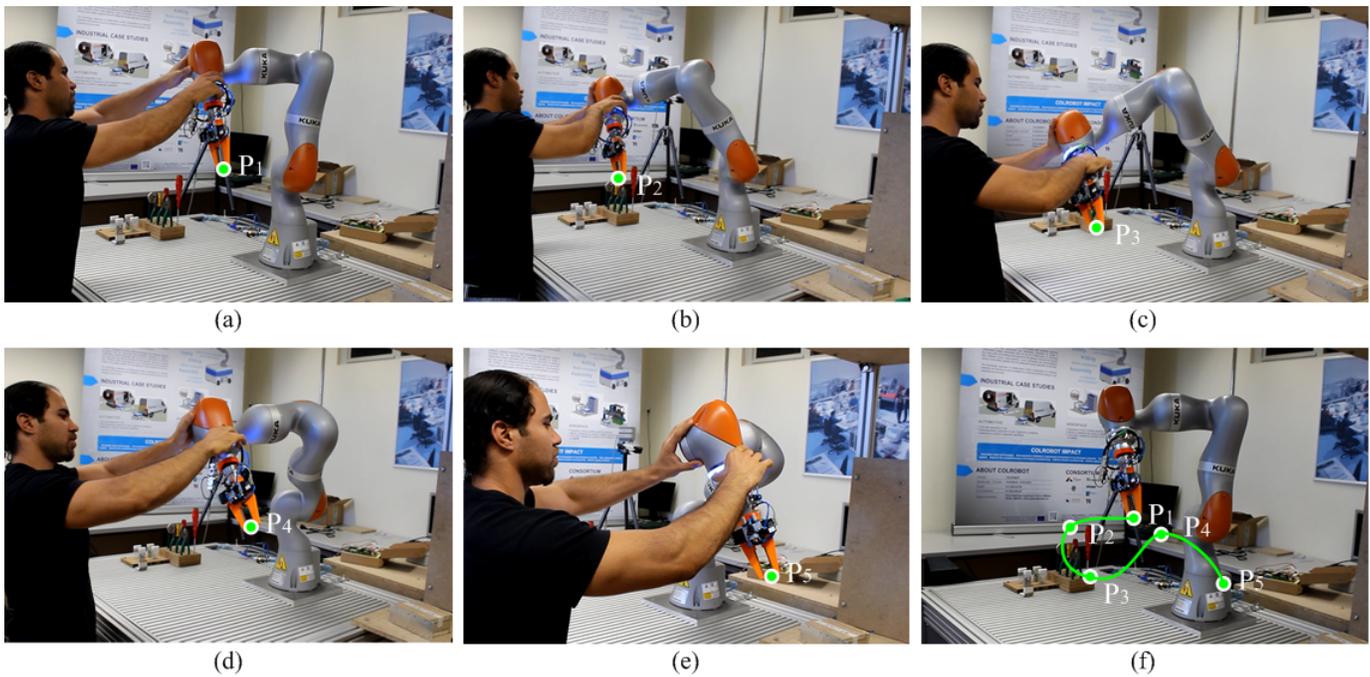

Fig. 2. Example 1: Hand-guiding teaching of five target points. (a) – (e) User teaching the robot path sequence with five points. (f) The robot automatically moves through the taught points.

The video that accompany this article show the three application examples https://youtu.be/nCIBYZ42uJw.

### A. Example 1: hand-guiding and teaching

Kinesthetic teaching is one of the main functionalities of collaborative robots. In this example (file `HandTeachingExample.m`) the user can teach the robot by hand-guiding it through a set of frames, five points in this specific example, Figure 2. When running the example file, the hand-guiding functionality is activated (`startHandGuiding`), so that the user is able to hand-guide the robot to a desired location. Once reached a target pose, in order to capture and save it the user has to click continuously the green button until the blue LED light starts to flash. This operation can be repeated to define other target poses. The taught poses are saved in MATLAB files allowing the automation of the process and its application to develop other robot applications with different sequencing if desired. The process is illustrated in the video.

### B. Example 2: Drawing a rectangle

The robot produces the drawing of a rectangle (a x b) on a white box with a pen mounted on the flange of the manipulator. The TCP/IP communication between KST and robot controller is initialized and the four points defining the vertices of the rectangle are sent to the robot using the point-to-point motion function `movePTPLineEEF`. The example code is in the MATLAB file `kuka0_move_sequare.m`. Figure 3 shows a snapshot of the robot drawing the rectangle.

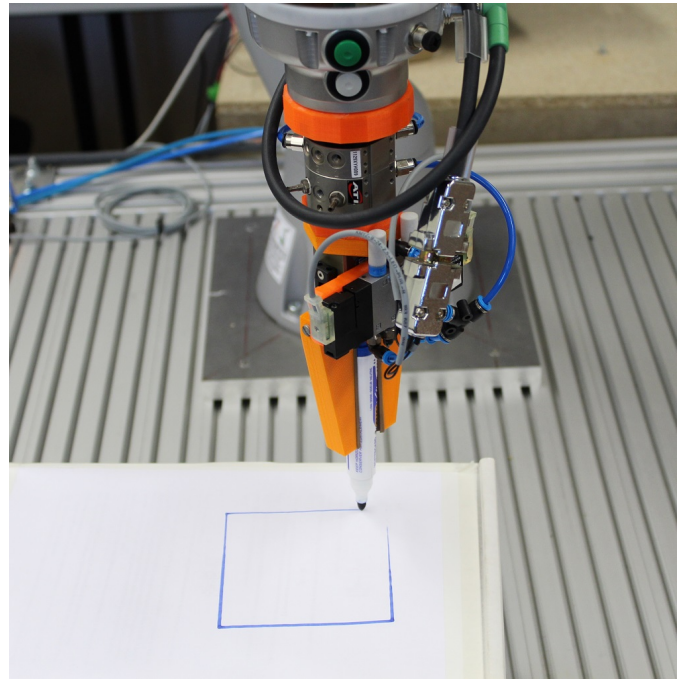

Fig. 3. Example 2: Robot drawing a rectangle.

### C. Example 3: Human-robot collision avoidance

The ability to have humans and robots working side-by-side and sharing the space is critical for the success of collaborative robots. In this example we propose human-robot collision avoidance based on the famous potential fields method [9]. In this example, the robot is in a home position (target) and when the human (obstacle) approaches the robot smoothly moves to avoid collision. The human pose in space is captured using a magnetic tracking sensor attached to the human hand.

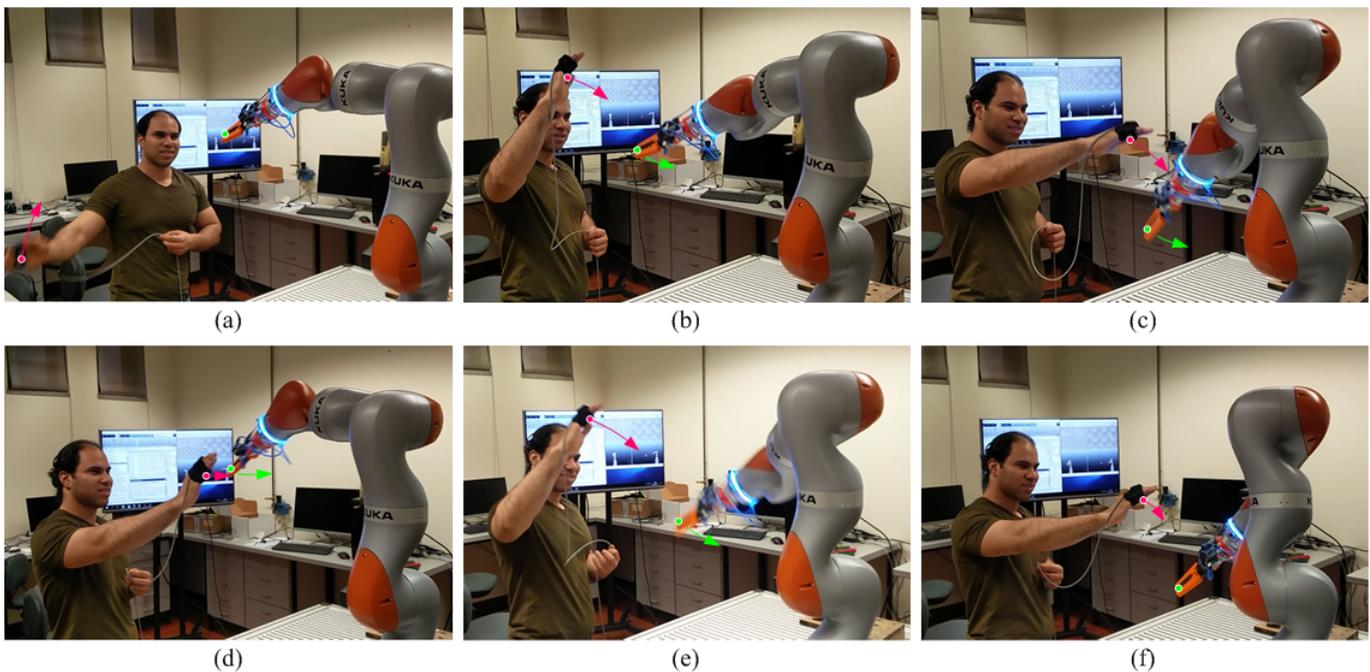

Fig. 4. Example 3: Human-robot collision avoidance. (a) – (e) The human co-worker tries to touch the robot but the robot reacts by moving to avoid collision. (f) Even with the human hand approaching the robot is not moving because its working area is limited by the table top.

The collision avoidance control algorithms were implemented in MATLAB allowing to speed up the implementation of advanced mathematical calculations (complex control algorithms, matrix manipulation, signal processing). Another advantage of using an external computer is related with the ease interfacing with external sensors to capture the human pose and the required computational power to run such algorithms and on-line generate robot motion to avoid collisions. The amount of calculations (we used an external computer with eight cores) required for the control algorithm renders the use of the robot controller alone unfeasible. The KST online updates the collision avoidance robot motion in the robot controller at a frequency of 275 Hz. The robot motion was controlled in joint space using the real-time control functions provided in the KST (`realTime_startDirectServoJoints` and `sendJointsPositions`). Figure 4 shows snapshots of the robot avoiding collisions with the human.

## IV. Conclusion

We presented the MATLAB KUKA Sunrise Toolbox (KST) to interface KUKA Sunrise.OS controller that equips the well-known KUKA LBR iiwa collaborative manipulator. The Toolbox runs on an external computer connected with the KUKA controller via TCP/IP. The KST functionalities for networking, real-time control, point-to-point motion, setters and getters of parameters and physical interaction, demonstrated reliability, versatility and ease of use. This performance was successfully validated in three application examples.

## V. Acknowledgment

This research was partially supported by Portugal 2020 project DM4Manufacturing POCI-01-0145-FEDER-016418 by UE/FEDER through the program COMPETE2020, and the Portuguese Foundation for Science and Technology (FCT) SFRH/BD/131091/2017.